\documentclass[journal,twoside,web]{ieeecolor}
\usepackage{caption}
\usepackage{booktabs}
\usepackage{tmi}
\usepackage{dblfloatfix}
\usepackage{array}
\usepackage[symbol]{footmisc}
\usepackage{caption}
\usepackage{subcaption}
\usepackage{dirtytalk}

\usepackage{cite}
\usepackage{lipsum}
\usepackage{hyperref}
\usepackage{amsmath,amssymb,amsfonts}
\usepackage{algorithmic}
\usepackage{graphicx}
\usepackage{multirow}
\usepackage{textcomp}
\usepackage{placeins}
\usepackage{fp}
\usepackage{xcolor}

\newcommand{\api}{AP\textsubscript{i}}
\newcommand{\app}{AP\textsubscript{p}}
\newcommand{\dice}{$\lceil$Dice$\rceil$}
\newcommand{\row}[5]{}  

\newcommand{\tsigp}[1]{
  \FPifgt{#1}{0.01}%
    \tiny{\textcolor{green}{+#1}}%
  \else%
    \tiny{\textcolor{gray}{+#1}}%
  \fi%
}
\newcommand{\tsign}[1]{
  \FPifgt{#1}{0.01}%
    \tiny{\textcolor{red}{–#1}}%
  \else%
    \tiny{\textcolor{gray}{–#1}}%
  \fi%
}

\newcommand{\tp}[1]{\tiny{\textcolor{gray}{+#1}}}

\newcommand{\tn}[1]{\tiny{\textcolor{gray}{–#1}}}

\newcommand{\tu}[1]{\tiny{\textcolor{gray}{+.00}}} 

\def\BibTeX{{\rm B\kern-.05em{\sc i\kern-.025em b}\kern-.08em
    T\kern-.1667em\lower.7ex\hbox{E}\kern-.125emX}}
\begin{document}

\title{Unsupervised Pathology Detection: A Deep Dive Into the State of the Art}
\author{
    \thanks{This article has been accepted for publication in IEEE Transactions on Medical Imaging. This is the author's version which has not been fully edited and
content may change prior to final publication. Citation information: DOI 10.1109/TMI.2023.3298093}
    Ioannis Lagogiannis, Felix Meissen, Georgios Kaissis and Daniel Rueckert, \IEEEmembership{Fellow, IEEE}
    \thanks{© 2023 IEEE.  Personal use of this material is permitted.  Permission from IEEE must be obtained for all other uses, in any current or future media, including reprinting/republishing this material for advertising or promotional purposes, creating new collective works, for resale or redistribution to servers or lists, or reuse of any copyrighted component of this work in other works.}
    \thanks{
    Ioannis Lagogiannis (i.lagogiannis@tum.de), Felix Meissen (felix.meissen@tum.de), Georgios Kaissis (g.kaissis@tum.de), and Daniel Rueckert (daniel.rueckert@tum.de) are  with the Chair for AI in Medicine and Healthcare (I31), School of Computation, Information and Technology, TU Munich, 85748 Garching and Klinikum rechts der Isar, 81675 München, Germany
    }
    \thanks{Georgios Kaissis, and Daniel Rueckert are with the Biomedical Image Analysis Group, Imperial College London, London SW7 2AZ, UK}
    \thanks{Georgios Kaissis is with the group for Reliable AI, Institute for Machine Learning in Biomedical Imaging, Helmholtz Zentrum München, Germany}
    \thanks{Ioannis Lagogiannis and Felix Meissen contributed equally.}
}

\maketitle

\begin{abstract}
Deep unsupervised approaches are gathering increased attention for applications such as pathology detection and segmentation in medical images since they promise to alleviate the need for large labeled datasets and are more generalizable than their supervised counterparts in detecting any kind of rare pathology.
As the Unsupervised Anomaly Detection (UAD) literature continuously grows and new paradigms emerge, it is vital to continuously evaluate and benchmark new methods in a common framework, in order to reassess the state-of-the-art (SOTA) and identify promising research directions.
To this end, we evaluate a diverse selection of cutting-edge UAD methods on multiple medical datasets, comparing them against the established SOTA in UAD for brain MRI.
Our experiments demonstrate that newly developed feature-modeling methods from the industrial and medical literature achieve increased performance compared to previous work and set the new SOTA in a variety of modalities and datasets. Additionally, we show that such methods are capable of benefiting from recently developed self-supervised pre-training algorithms, further increasing their performance. Finally, we perform a series of experiments in order to gain further insights into some unique characteristics of selected models and datasets. Our code can be found under \url{https://github.com/iolag/UPD_study/}.
\end{abstract}
\begin{IEEEkeywords}
Unsupervised, Anomaly, Detection, Segmentation, Medical, Comparative, Generative, Image-reconstruction, Feature-modeling, Self-supervised, pre-training
\end{IEEEkeywords}

\section{Introduction}\label{sec:introduction}
\IEEEPARstart{F}{rom} routine check-ups to the detection and treatment of brain tumors, pathology detection from medical images is an indispensable part of the clinical diagnosis and treatment workflow. While generally performed manually by clinical experts (e.g. radiologists), an ever-increasing demand for radiological assessments in modern healthcare systems has prompted researchers to focus on developing algorithmic solutions that can assist in clinical diagnosis. 
Anomaly Detection (AD) can be described as an outlier detection problem, where the aim is to discriminate between in- and out-of-distribution samples of a normative distribution.
In the context of Pathology Detection (PD) for medical diagnosis, the \say{normal} distribution constitutes healthy samples and cases containing pathologies can be detected as outliers.
While PD can be tackled with supervised learning strategies, the main concerns here are twofold.
Firstly, they require vast amounts of image-level labels or pixel-level segmentations from medical experts, but these are scarce and costly to obtain, posing practical limitations for training such models.
Additionally, the morphological variability of pathologies is large and rare anomalies are likely underrepresented (or not included at all) in the datasets, posing problems for supervised methods.
On the contrary, Unsupervised Pathology Detection (UPD) methods leverage exclusively healthy samples during training in order to model the normal anatomy distribution. Given that a significant part of acquired medical images, i.e. from routine or preventive check-ups, is clinically unremarkable, the unsupervised setting theoretically reveals a large amount of data to train UPD models.
Therefore, there has been a recent surge in the development of UPD methods \cite{Baur_anovaegan,baur2021autoencoders,fanogan,anogan,zimmerer-vae,gmvae,AE}. 
Unfortunately, the lack of publicly available benchmark datasets for medical UPD forces practitioners to develop and evaluate their models on various private and public datasets, hindering comparability and the development of best practices in this field. The problem is further exacerbated by the concurrent development of UAD methods in other areas such as industrial inspection \cite{MVTecDataset}.

To this end, the following work presents a thorough comparison of the most common current UAD paradigms, applied to detecting pathologies in medical images.
Specifically, we evaluate the anomaly detection and localization performance of 13 UPD methods on 4 different medical datasets with different characteristics.
To the best of our knowledge, this is the most comprehensive study so far, covering all important paradigms on a representative set of modalities.

\section{Related Work}


In this section, we briefly describe recent advances in Deep Unsupervised Anomaly Detection and Localization. We build our categorization upon the work of Jie \textit{et al.} \cite{survey2} and identify four groups: \textit{image-reconstruction} \cite{baur2021autoencoders,AE,fanogan,zimmerer-vae,Baur_anovaegan,gmvae,constrainedAE}, \textit{feature-modeling} \cite{DFR,FAE,stud-teach,RD,uninformed_students,cflow,fastflow,mvg_mahalanobis,padim,panda_knn,spade}, \textit{attention-based} \cite{cavga,expvae,AMCons}, and \textit{self-supervised anomaly detection} \cite{pii,cutpaste,DAE, patchSVDD, autoseg} methods.

\subsection{Image-reconstruction Methods}
The prevailing \textit{modus operandi} in UPD uses image-reconstruction models to model normality and detect deviations from it. 
Vanilla \cite{AE} or Variational Autoencoders \cite{zimmerer-vae}, Generative Adversarial Networks \cite{fanogan,anogan}, or combinations and variations of these frameworks \cite{Baur_anovaegan,gmvae,constrainedAE} have been explored.
During inference, such models use the residual $\mathbf{r =\left| x - \hat{x}\right|}$ between the input image $\mathbf{{x}}$ and its reconstruction $\mathbf{\hat{x}}$ to generate a saliency map.
As a novel extension of the reconstruction paradigm, the restoration-approach \cite{gmvae}, an input image is iteratively updated until its anomalous regions are replaced with quasi-healthy ones.
Baur \textit{et al.} \cite{baur2021autoencoders} performed a systematic evaluation of a collection of image-reconstruction UPD algorithms in the context of brain MR imaging. We refer the interested reader to their work, for a thorough explanation of the above concepts.

\subsection{Feature-modeling Methods}
Newly developed algorithms, mainly from the field of industrial inspection with UAD, have started deviating from the image-reconstruction norm. Instead of working on the image directly, \textit{feature-modeling} methods leverage frozen, pre-trained encoders to first transform each input sample to an alternative, semantically-rich representation, which they proceed to manipulate and model with a variety of techniques in order to perform anomaly detection. 
A popular research direction adopts the student-teacher learning paradigm \cite{stud-teach,RD,uninformed_students}, distilling knowledge from the pre-trained encoder to a student network. By enforcing similarity in student and teacher activations during training, representation discrepancies can reveal anomalies during inference.
Feature-modeling generative approaches are also gaining traction, with researchers using normalizing flow (NFLOW) networks to model normal embeddings in an attempt to estimate exact likelihoods \cite{cflow,fastflow}.
Even simpler statistical baselines that do not require gradient optimization have achieved SOTA detection and segmentation performance. Such methods act directly on pooled features and often employ Multivariate Gaussian \cite{mvg_mahalanobis,padim} or K-nearest neighbor \cite{panda_knn,spade} modeling to capture normality. Impressive in their simplicity and detection capabilities, these approaches often exhibit long inference times, limiting their practical applications.
Lastly, the image-reconstruction approach has been successfully applied to feature-embeddings \cite{FAE}. 

\subsection{Attention-based Methods}
When learning normality with a machine learning model, using attention maps from layer-activations or the gradient of the normality formulation is a natural fit to extract localization information from the trained model. This principle was applied by Zimmerer \textit{et al.} in one of their model variants in \cite{zimmerer-vae}, and by Liu \textit{et al.}, Venkataramanan \textit{et al.}, and Silva-Rodriguez \textit{et al.} \cite{expvae,cavga,AMCons} in the form of GradCAM \cite{gradcam} maps.

\subsection{Self-supervised Anomaly Detection Methods}

Self-supervised methods are gaining popularity in both the medical and industrial inspection fields, with researchers designing and performing pre-text tasks on normal data to perform UAD.
One popular research direction utilizes pre-text tasks in order to initially perform representation learning, and then proceeds to model the representation distribution of normal instances\cite{patchSVDD, cutpaste}. Such methods detect anomalies as outliers from the modeled distribution, premised on the notion that the trained model can generalize and effectively map anomalous inputs in out-of-distribution areas of the representation space.
Another approach involves the application of synthetic anomalies on otherwise normal training data and the adoption of common supervised techniques to explicitly \cite{pii,autoseg} or implicitly \cite{DAE} localize them.


\subsection{Self-supervised Pre-training Methods}
Apart from anomaly detection, self-supervised methods have predominantly been used to 
learn useful representations by performing pretext tasks on unlabeled data \cite{simclr}. Such tasks include but are not limited to detecting geometric transformations \cite{self-sup-geom,self-sup-rotation} and contrastive learning tasks \cite{simclr,moco}. The learned, domain-specific representations can be used in conjunction with detection methods in UAD pipelines or any other downstream task.

\section{Motivation and Contribution}
The primary objective of this work is to gain a clear and up-to-date picture of the state of Unsupervised Pathology Detection. To this end, we conducted a comparative study of state-of-the-art UAD approaches on multiple medical imaging modalities.
Prompted by the limitations of image-reconstruction methods \cite{meissen_flair,meissen_pitfalls} and the latest developments in industrial UAD, especially regarding the overwhelming popularity of the feature-modeling paradigm, we included a group of unique feature-modeling approaches next to methods trained with self-supervised learning, recent attention-based methods, and the SOTA in image-reconstruction brain MR UPD as established in the recent work of Baur \textit{et al.} \cite{baur2021autoencoders}.
As industrial UAD methods are mainly developed for defect detection applications and evaluated on MVTec AD \cite{MVTecDataset}, a dataset constituting an average of 242 low variation RGB samples per class, we are also interested in whether high performance on it generally translates to the much different Medical Imaging domain.
Furthermore, we studied the effects of self-supervised pre-training on all methods. As feature-modeling approaches make use of pre-trained, frozen encoders, the ability of methods to take advantage of SOTA pre-training schemes is especially relevant to our investigation.
We chose to evaluate the selected methods on 4 publicly available datasets of 3 different medical imaging modalities. By studying how selected approaches fare against this diverse selection of datasets with different characteristics regarding image and anomaly appearance, normal anatomy variation as well as the number of available samples, we can get a fuller picture of each method's capabilities, and also gain valuable insights into the modalities themselves.
Finally, we performed a time- and space-complexity analysis, and examined how lesion size and intensity as well as the number of available train samples affects UPD performance, gaining valuable insights into selected methods and datasets.
\section{Reviewed Methods}
In this section, we briefly describe the methods included in our study.

\subsection{Image-reconstruction Models}
To allow us to set our results into perspective with those reported in Baur \textit{et al.} \cite{baur2021autoencoders}, we selected the 3 best-performing methods from their study as representatives of the image-reconstruction paradigm.
The methods are the dense \textbf{VAE} \cite{kigma-vae}, the \textbf{f-AnoGAN} \cite{fanogan}, and the VAE with iterative restoration \cite{gmvae}, which we will denote as \textbf{r-VAE} in the rest of this manuscript.
We also included one of the best performers from a  transformer-based group of methods recently developed by Ghorbel \textit{et al.}\cite{tae}, namely the Hierarchical Transformer Autoencoder with Skip connections (\textbf{H-TAE-S}). As a novel non-CNN-based method, it forms an Autoencoder framework using transformer blocks, inspired by the Swin-Unet architecture\cite{swin-unet}.
All of the above methods are reconstruction-based, meaning they are trained to reconstruct a "healthy" version of the input image and use the residual between the input- and the reconstructed "healthy" image to detect anomalies.

\subsection{Feature-modeling Models}
When deciding on SOTA methods that were developed in the context of industrial inspection, our criteria were not limited to reported performance results on the MVTec dataset. As mentioned above, one of our main objectives in this study is to examine whether SOTA performance from the diverse industrial UAD literature can translate to the medical domain. Therefore, we incorporated approaches that employ a wide variety of learning paradigms and techniques to perform UAD. Still, feature extraction by a pre-trained and frozen encoder network is the common characteristic of all selected methods.
Our first choice, Reverse Distillation (\textbf{RD}) \cite{RD}, combines an AE architecture with a student-teacher knowledge distillation approach, forcing the student decoder to recreate representations produced by the pre-trained and frozen teacher encoder. During inference and in the presence of anomalies, decoder and encoder activations cannot remain consistent, and cosine distance is used to measure dissimilarities between corresponding feature maps and localize the anomalies.
\textbf{CFLOW-AD} \cite{cflow} adopts a Normalizing Flow \cite{nflows-orig} framework to estimate, contrary to the rest of the selected methods, exact pixel-level likelihoods. During training, it extracts activation maps from three specific layers of a pre-trained and frozen encoder. It models normality on each layer independently, using three normalizing flow decoders. At test time, it aggregates the outputs from all three decoders to produce the final anomaly map.
For each training sample, \textbf{PaDiM} \cite{padim} makes a single forward pass through a pre-trained encoder network and creates a dataset-wide embedding volume. It then computes the mean and covariance of each feature vector to infer on test samples, using the Mahalanobis distance. While being a straightforward yet powerful approach that requires no gradient optimization, it suffers from extreme memory demands and slow inference times.
Lastly, we included Structural Feature-Autoencoders (\textbf{FAE}) \cite{FAE} as an application of the reconstruction-based paradigm on feature embeddings. It projects the image into a higher-dimensional feature space using a feature-extraction network and uses multi-channel residuals between model input and output to localize anomalies in this space. FAE uses an Autoencoder architecture and the Structural Similarity Index Measure (SSIM) \cite{SSIM} as a target for optimization.

\subsection{Attention-based Methods}
As the first representative for this category, we evaluate \textbf{ExpVAE} \cite{expvae} on our selected medical datasets. ExpVAE is a Variational Autoencoder, equipped with a GradCAM mechanism applied on the the KL-divergence to generate attention maps that can be used instead of residuals for anomaly localization.
Our second representative, \textbf{AMCons} \cite{AMCons}, extends on ExpVAE by using only the feature-activations of a certain layer instead of the whole GradCAM formulation and by adding a novel regularization term to its objective function.

\subsection{Self-supervised Anomaly Detection Methods}
As a representative of self-supervised anomaly detection methods, we used \textbf{PII} \cite{pii}. During training, the method first samples a patch from a random healthy image and seamlessly blends it into a random location of the input using Poisson image editing \cite{poisson} and a random interpolation factor. It then trains end-to-end by learning to localize the altered patch and predicting the interpolation factor. The trained model is later used to directly localize anomalies.
Our second choice in this category, the Denoising Autoencoder (\textbf{DAE}) \cite{DAE}, generates synthetic anomalous samples by adding Gaussian noise to each input and trains a model to remove that noise. During inference, anomalous regions are detected via the pixel-wise residual between the input image and the model output, analogous to reconstruction-based methods.
Lastly, we select \textbf{CutPaste} \cite{cutpaste}. This method trains two separate encoders in order to learn compact representations of input images (for image-level detection) or patches (for pixel-level localisation) using a 3-way classification task between each input and two versions with artificial anomalies. 
It then performs statistical modeling on the representations of normal samples and anomalies are detected as outliers from the normal representation distribution.
Note that while DAE computes anomaly maps by extracting residuals like the image-reconstruction methods and CutPaste by modeling deep representations similar to feature-modeling approaches, both are trained in a self-supervised way using synthetic anomalies.


\begin{figure*}[htb]
\centerline{\includegraphics[width=\textwidth]{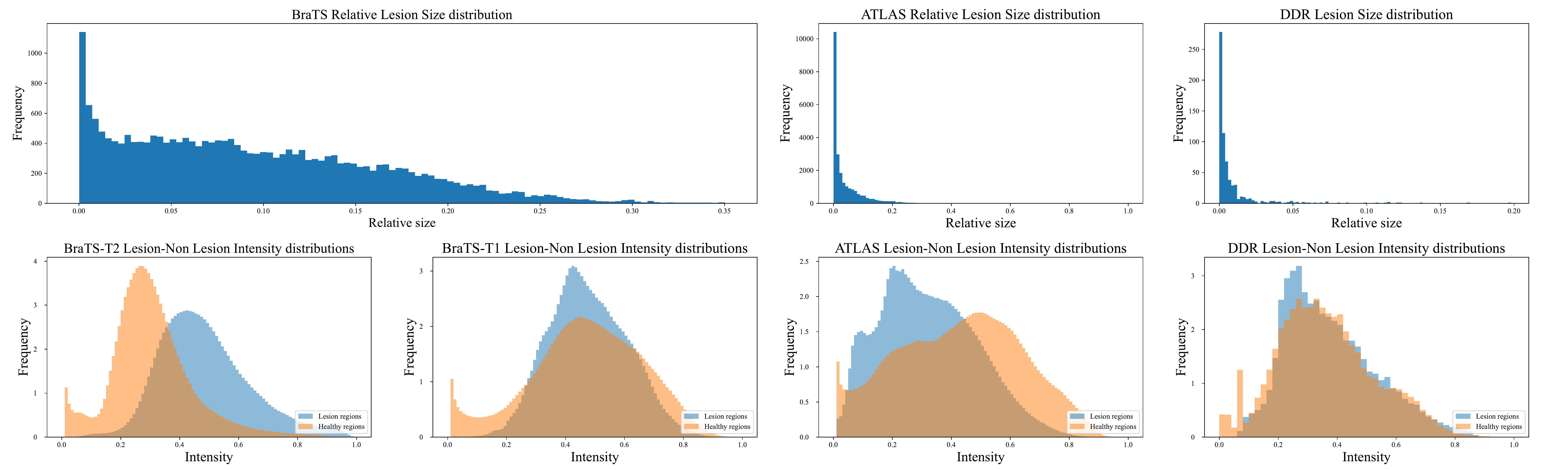}}
\caption{Statistics of the datasets used for pixel-wise evaluation in this study: BraTS, ATLAS, and DDR. First row: Distributions of lesion sizes relative to the brain area (BraTS and ATLAS) or the total image (DDR). Second row: Distributions of mean lesion intensities for normal and anomalous regions.}
\label{fig:stats}
\end{figure*}

\section{Experimental Setup}

\begin{table}[t]
\centering
\caption{Number of images used for training, validating, and testing each dataset.}
\label{tab:splits}
\resizebox{\linewidth}{!}{
\begin{tabular}{lcccc} 
\toprule
Modality & Training & Validation & Testing & Test Val. \\ 
\cmidrule{1-5}
CamCAN & 85752 & 4514 & - & - \\
ATLAS & - & - & 70674 & 7920 \\
BraTS & - & - & 46557 & 5150  \\
CheXpert & 4267 & 255 & 720 & 80 \\
DDR & 5233 & 276 & 1340 & 150 \\
\cmidrule[\heavyrulewidth]{1-5}
\end{tabular}}
\end{table}

\subsection{Modalities and Datasets}\label{sec:datasets}


Unlike its industrial inspection counterpart, the lack of benchmark datasets specifically designed for the task remains a pertinent issue in UPD, resulting in a plethora of different public and private datasets used for the evaluation of individual methods. This variety, however, hinders the comparability between the proposed methods and prevents us from drawing general conclusions about different approaches.
To tackle this issue, we introduce a selection of multiple publicly available and commonly used datasets of different modalities that allow for direct evaluation of the methods regarding different data characteristics and even allow future methods to assess their performance regarding these characteristics.
Unlike \cite{baur2021autoencoders}, we decided not to use FLAIR MR scans, as most anomalies there appear hyperintense, resulting in models only being evaluated regarding their "white object" detection capabilities \cite{meissen_flair,meissen_pitfalls}.
Instead, we opted to train and test on T1- and T2-weighted brain scans (\textbf{brain-MRI}).
We also expanded our evaluation with two additional modalities, namely Chest X-ray (\textbf{CXR}) and Retinal Fundus Photography (\textbf{RF}), in an attempt to establish a more comprehensive benchmark for UPD.
\smallskip\newline
\textbf{Brain-MRI}
We used the Cambridge Centre for Ageing and Neuroscience dataset (CamCAN) \cite{camcan} as our source of lesion-free scans for all Brain-MRI datasets. It contains T1 and T2-weighted brain scans of 652 healthy adult subjects.
The Anatomical Tracings of Lesions After Stroke (ATLAS) dataset \cite{atlas} provided us with 220 T1-weighted scans of stroke patients, which we use for evaluation.
We also used the 2020 version of the Multimodal Brain Tumor Segmentation (BraTS) Challenge dataset for evaluation \cite{brats1,brats2,brats3}. It consists of 371 multiple-sequence MR scans of patients suffering from high-grade glioblastomas or low-grade gliomas.
We denote the sets of T1- and T2-weighted MR scans from BraTS as BraTS-T1 and BraTS-T2 respectively in the rest of this manuscript. Expert lesion segmentations are provided for both ATLAS and BraTS, allowing for pixel-level evaluation.
After pre-processing (c.f. Section \ref{sec:preprocessing}), the BraTS test set contains 22\,422 anomal and 24\,135 normal axial slices, and the ATLAS test set 23\,989 anomal and 56\,590 normal ones.
Note that due to the domain shift between the training- and test-datasets, the anomaly detection models used in this study are expected to output higher anomaly scores for all test images.
This problem is partly mitigated because both normal and anomal images in the test sets are from the same datasets.
\smallskip\newline
\textbf{CXR}
to evaluate our selected models on the challenging CXR modality, we used CheXpert \cite{chexpert}, a large publicly available dataset containing 224\,316 chest radiographs. Samples are labeled as positive, negative, or uncertain for 14 observations, 12 of which are pathology related. The remaining two regard the presence of support devices or the absence of all pathologies, labeled as "no finding". We use the last observation for our "healthy" subset but we do so with caution as the label only implies the absence of the 12 pre-defined pathologies of interest, a common limitation in CXR datasets.
We limited the task by using images annotated with exactly one out of three of the most common observations, namely "pleural effusion", "opacity" or "enlarged cardiomediastinal contour".
We avoided possible confounders and biases by discarding scans with support devices, using only radiographs of anteroposterior view, and by enforcing an equal number of samples of male and female patients on all consequent subsets.
We created three test sets comprising 400 "no finding" and 400 cases labeled as positive for one of the aforementioned observations, which left us with a train set of 4492 "no finding" images.
Note that for CheXpert, no expert segmentations are available so only sample-wise evaluation is possible.
This evaluation is interesting nevertheless, as pathological findings in Chest X-ray images are very subtle, and detecting them automatically has high clinical value.
\smallskip\newline
\textbf{RF}
Finally, we evaluated the compared methods on the DDR dataset \cite{DDR}, which provides 6243 fundus photographs of healthy individuals and 745 samples of positive Diabetic Retinopathy (DR) cases with corresponding ground truth segmentations for the four types of lesions associated with DR. We used 5510 healthy samples for training and created a test set out of 745 healthy and 745 DR-positive samples. Since UPD methods are agnostic to the types of anomalies they detect, for each sample we combined all four provided binary masks for the different types of lesions into a single reference segmentation to evaluate. 
\smallskip
\newline
The above datasets have very different characteristics in both the distribution of normal samples and in their anomalies of interest. While the brain-MRI and RF datasets have low intra-sample variance, the organs seen in chest radiographs exhibit a larger variety among different patients. Further, anomalies are more localized in brain-MR and RF images, whereas they are more diffuse in CXR. The DDR dataset is a special case not only because it contains by far the smallest lesions (see Fig. \ref{fig:stats}), but also because it is the only dataset with RGB images instead of grayscale ones. The plots in Fig. \ref{fig:stats} reveal that anomalies tend to be hypo-intense in the ATLAS dataset, hyper-intense in BraTS-T2, and neither of both in BraTS-T1 and the RF images.

\subsection{Pre-processing} \label{sec:preprocessing}
All brain-MRI datasets were registered to the SRI24 ATLAS \cite{SRI}, skull-stripped, and sliced axially into 2D images. For both training and evaluation, we discarded slices with no brain pixels in them and split each subset volume-wise to eliminate patient overlap between the subsets and maintain complete brain scans. During the image-wise evaluation, a slice with any number of positive pixels in its ground truth mask was considered anomalous.
Across all modalities, samples were resized to $128 \times 128$ pixels and scaled into the range [0, 1]. For RF, the only RGB dataset, we opted for scaling into [-1, 1] as we noticed improved performance across the board in early experiments.
For all datasets, we kept 10\% of test samples for validation and 5\% of normal samples for monitoring the loss during training. All splits are listed in Table \ref{tab:splits}.

\subsection{Model architectures and Hyper-parameters}
In contrast to Baur \textit{et al.} \cite{baur2021autoencoders}, we did not use a unified architecture, as such practice does not allow for a fair comparison given the diversity of our selected methods. Instead, we used the original architectures and hyper-parameters, optimizing for the datasets at hand when necessary. We still enforced commonality in other parts of the pipeline, such as the pre- and post-processing and in a common pre-trained backbone choice when applicable.  
We refer to the respective papers for detailed descriptions of the architectures.

\subsection{Residual extraction and Post-processing}
For residual-based methods (VAE, r-VAE, f-AnoGAN, FAE, 
and DAE in this study), anomaly maps are typically extracted from the pixel-wise absolute error $\mathbf{r =\left| x - \hat{x}\right|}$ between the input image $\mathbf{{x}}$ and its reconstruction $\mathbf{\hat{x}}$.
Instead, we opted to use the Structural Similarity Index (SSIM) \cite{SSIM} (as in \cite{ssimae} and \cite{FAE}) as a more powerful measure of dissimilarity that takes not only differences in luminance, but also texture and structure into account.
This change increased the performance of all selected models it was applied to.
Please refer to \cite{ssimae} for an analysis of this choice.
For anomaly maps of MR inputs, we discarded background pixels as possible sources of anomalies. Aside from that, no other post-processing steps were taken to ensure a fair comparison among methods.
For sample-level scoring, a common practice in the literature is taking the mean over all pixel values in the anomaly map \cite{zimmerer-vae,fanogan,pii,FAE}.
For brain-MRI and RF, we instead used the value of the maximally activated pixel as the sample-level score (as in \cite{RD,cflow}), since it should be independent of the anomaly size.
Further, the anomalies in the DDR dataset are so small that even correct localization would only lead to a slight increase in the averaged anomaly map.
On the other hand, for CXR images that are usually littered with markers and other artifacts that are expected to cause high activations, using the mean proved a more robust choice in early experiments.

\subsection{Metrics}
To evaluate image-level detection performance, we used the area under the receiver operating characteristic curve (AUROC) and image-wise average precision (AP\textsubscript{i}), which is equivalent to the area under the precision-recall curve.
Both metrics allow for model assessment without the need to choose an Operating Point (OP).
For the evaluation of pixel-level performance, we used pixel-wise average precision (AP\textsubscript{p}) and an estimate of the best possible S{\o}rensen-Dice index ($\lceil$Dice$\rceil$).
All metrics were computed dataset-wise.


\section{Results and Discussion} \label{sec:results}

\renewcommand{\row}[5]{ #1 & #4 && #3 && #2 && #5}
\begin{table*}[htb]
    \centering
    \caption{Detection and localization results of the image-reconstruction (\textbf{IR}), feature-modeling (\textbf{FS}), attention-based (\textbf{AB}), and self-supervised anomaly detection (\textbf{S-S}) methods on the brain-MRI and RF datasets. Best scores are bold.}
    \label{tab:main}
    \resizebox{\linewidth}{!}{
    \begin{tabular}{clccccccccccccccccccc}
        \toprule
        & \row{}{\multicolumn{4}{c}{\textbf{ATLAS}}}{\multicolumn{4}{c}{\textbf{BraTS-T1}}}{\multicolumn{4}{c}{\textbf{BraTS-T2}}}{ \multicolumn{4}{c}{\textbf{DDR}}}\\
        \cmidrule(r{4pt}){1-2}\cmidrule{3-6}\cmidrule{8-11}\cmidrule{13-16}\cmidrule{18-21}
        & & \multicolumn{2}{c}{image-level} & \multicolumn{2}{c}{pixel-level} && \multicolumn{2}{c}{image-level} & \multicolumn{2}{c}{pixel-level} && \multicolumn{2}{c}{image-level} & \multicolumn{2}{c}{pixel-level} && \multicolumn{2}{c}{image-level} & \multicolumn{2}{c}{pixel-level} \\[1mm]
        & Method & \api & AUROC & \app & \dice && \api & AUROC & \app & \dice && \api & AUROC & \app & \dice && \api & AUROC & \app & \dice \\
        \cmidrule(r{4pt}){1-2}\cmidrule{3-6}\cmidrule{8-11}\cmidrule{13-16}\cmidrule{18-21}
        \parbox[t]{2mm}{\multirow{4}{*}{\rotatebox[origin=c]{90}{\textbf{IR}}}}
        & \row{VAE}{0.57 & 0.76 & \textbf{0.11} & 0.20}{0.64 & 0.70 & 0.13 & 0.19}{0.68 & 0.73 & 0.28 & 0.33}{0.48 & 0.48 & 0.02 & 0.05} \\
        & \row{r-VAE}{\textbf{0.60} & \textbf{0.78} & 0.09 & 0.17}{0.70 & 0.76 & 0.13 & 0.19}{0.75 & 0.77 & 0.36 & 0.40}{0.52 & 0.50 & 0.03 & 0.09} \\
        & \row{f-AnoGAN}{0.26 & 0.46 & 0.02 & 0.06}{0.48 & 0.53 & 0.06 & 0.12}{0.56 & 0.61 & 0.15 & 0.21}{0.44 & 0.45 & 0.01 & 0.01} \\
         & \row{H-TAE-S}{0.29 & 0.49 & 0.01 & 0.03}{0.54 & 0.57 & 0.06 & 0.12}{0.68 & 0.70 & 0.21 & 0.12}{0.51	& 0.51 & 0.01 & 0.01} \\
        \cmidrule(r{4pt}){1-2}\cmidrule{3-6}\cmidrule{8-11}\cmidrule{13-16}\cmidrule{18-21}
        \parbox[t]{2mm}{\multirow{4}{*}{\rotatebox[origin=c]{90}{\textbf{FM}}}}
        & \row{FAE}{0.50 & 0.73 & 0.08 & 0.18}{\textbf{0.86} & \textbf{0.85} & \textbf{0.42} & \textbf{0.45}}{\textbf{0.87} & \textbf{0.87} & \textbf{0.51} & \textbf{0.52}}{0.64 & 0.63 & 0.07 & 0.15} \\
        & \row{PaDiM}{0.34 & 0.56 & 0.05 & 0.13}{0.60 & 0.65 & 0.21 & 0.28}{0.66 & 0.68 & 0.34 & 0.38}{0.55 & 0.55 & 0.02 & 0.07} \\
        & \row{CFLOW-AD}{0.40 & 0.62 & 0.04 & 0.10}{0.65 & 0.69 & 0.16 & 0.24}{0.71 & 0.72 & 0.31 & 0.35}{0.51 & 0.51 & 0.03 & 0.08} \\
        & \row{RD}{0.55 & 0.77 & \textbf{0.11} & \textbf{0.22}}{0.81 & 0.83 & 0.36 & 0.42}{0.85 & 0.85 & 0.47 & 0.50}{\textbf{0.66} & \textbf{0.64} & \textbf{0.10} & \textbf{0.19}} \\
        \cmidrule(r{4pt}){1-2}\cmidrule{3-6}\cmidrule{8-11}\cmidrule{13-16}\cmidrule{18-21}
        \parbox[t]{2mm}{\multirow{2}{*}{\rotatebox[origin=c]{90}{\textbf{AB}}}}
        
        & \row{ExpVAE} {0.37 & 0.57 & 0.01 & 0.03}{0.56 & 0.56 & 0.07 & 0.13}{0.63 & 0.66 & 0.12 & 0.18}{0.53 & 0.54 & 0.004 & 0.01}\\
        & \row{AMCons}{0.32 & 0.53 & 0.01 & 0.03}{0.61 & 0.64 & 0.05 & 0.12}{0.78 & 0.78 & 0.35 & 0.40}{0.49 & 0.49 & 0.004 & 0.01} \\
        \cmidrule(r{4pt}){1-2}\cmidrule{3-6}\cmidrule{8-11}\cmidrule{13-16}\cmidrule{18-21}
        \parbox[t]{2mm}{\multirow{3}{*}{\rotatebox[origin=c]{90}{\textbf{S-S}}}}
        & \row{PII}{0.37 & 0.60 & 0.03 & 0.07}{0.54 & 0.64 & 0.13 & 0.22}{0.57 & 0.62 & 0.13 & 0.22}{0.62 & 0.63 & 0.01 & 0.01} \\
        & \row{DAE}{0.44 & 0.65 & 0.05 & 0.13}{0.70 & 0.74 & 0.13 & 0.20}{0.81 & 0.80 & 0.47 & 0.49}{0.54 & 0.55 & 0.01 & 0.03} \\
        & \row{CutPaste}{0.37 & 0.58 & 0.03 & 0.06}{0.61 & 0.65 & 0.07 & 0.13}{0.59 & 0.63 & 0.22 & 0.26}{0.64 & 0.60 & 0.02 & 0.06}\\
        \cmidrule(r{4pt}){1-2}\cmidrule{3-6}\cmidrule{8-11}\cmidrule{13-16}\cmidrule{18-21}
        & \row{Random}{0.30 & 0.50 & 0.02 & 0.03}{0.48 & 0.50 & 0.06 & 0.11}{0.48 & 0.50 & 0.06 & 0.11}{0.50 & 0.50 & 0.004 & 0.01} \\
        \bottomrule
    \end{tabular}}
\end{table*}

\subsection{Main Results} \label{subsec:main_results}
Tables \ref{tab:main} and \ref{tab:cxr} include evaluation results for all selected approaches and modalities. We report the mean over three runs of different seeds. 
For each dataset, we also include the performance a random classifier would achieve as a baseline value for AP. This allows us to set the results of the heavily unbalanced segmentation tasks into perspective. Below, we analyze and discuss these results.
\smallskip\newline
\textbf{BraTS-T2: Big, hyper-intense lesions.}
As shown in Fig. \ref{fig:stats}, BraTS-T2 is characterized by relatively large, rather hyper-intense lesions. This is a setting most similar to the FLAIR datasets in Baur \textit{et al.} \cite{baur2021autoencoders}, although the overlap between the intensities of normal and anomal regions is significantly bigger in T2 than in FLAIR.
We observe superior performance of RD and FAE compared to the rest of the architectures.
Direct runner-ups are DAE and AMCons. Note, however, that these methods were developed for tumor detection on BraTS. Other candidates from the attention-based and self-supervised groups are among the worst-performing models along with f-AnoGAN and H-TAE-S.
r-VAE is the best-performing image-reconstruction method, surpassing PaDiM and CFLOW-AD.
\smallskip\newline
\textbf{BraTS-T1: Effects of limited contrast.}
The anomalies in BraTS-T1 are as large as in BraTS-T2, but generally, neither hyper- nor hypo-intense (see Figure \ref{fig:stats}), making them harder to detect.
As a result, a drop in performance (especially for segmentation) can be noticed for all methods, compared to the T2 case.
The least affected and still the best-performing models for this modality are the feature-modeling methods (especially FAE and RD), suggesting that these methods either benefit from bigger lesions and/or are more capable of capturing morphological and texture discrepancies.
Note, however, that FAE was developed for good performance on this dataset and also by multiple authors of this manuscript.
In the image-reconstruction group, the segmentation performance of all methods is more than halved and the scores of f-AnoGAN and H-TAE-S are only slightly above random, confirming the findings of \cite{meissen_pitfalls} that this class of methods is unable to localize anomalies that are neither hyper- nor hypo-intense.
Surprisingly, AMCons fails dramatically on BraTS-T1, a behavior that we will also observe in all later datasets. Please refer to Section \ref{sec:qualitative_results} for further analysis of this phenomenon.
In the self-supervised category, PII and CutPaste remain on a low level and DAE drops dramatically.
The answers to the large performance drop of DAE can be found in the method itself. It was trained to remove upsampled Gaussian noise, i.e. hyper- or hypo-intense blobs, from the images.
If the anomalies of interest are neither, the model has little chance of finding them.
\smallskip\newline
\textbf{ATLAS: Smaller and hypo-intense.}
Compared to BraTS-T1 and -T2, the lesions in the ATLAS dataset are generally smaller and appear hypo-intense to the normal tissue (see Figure \ref{fig:stats}). Here, r-VAE and VAE slightly outperform RD in image-level detection, while VAE and RD are the strongest in segmentation.
Compared to their results on BraTS-T2, PaDiM, and CFLOW-AD clearly underperform here, revealing a potential weakness with small lesions. We hypothesize this to be rooted in the spatially reduced activation maps used by feature-modeling methods.
The performance of the self-supervised methods drops the most, suggesting that their synthetic anomalies and pre-text tasks do not generalize well to smaller, hypo-intense anomalies. For DAE this is not surprising, given that the upsampling factor of the added Gaussian noise was optimized for best performance on BraTS.
In the attention-based group, ExpVAE and AMCons once again showcase little to no performance.
While the relative performance of most models compared to a random classifier dropped from BraTS-T1 to ATLAS, it increased for VAE and r-VAE. Thus, lesion size does not seem to have a significant influence on these two models. A further discussion on this follows in Section \ref{sec:influence_of_size_and_intensity}.
This, however, does not hold for the other two image-reconstruction methods.
\smallskip\newline
\textbf{DDR.}
As DDR has by far the smallest anomalies that even SOTA supervised methods struggle to localize \cite{DDR}, low performance by all methods was expected. Surprisingly, FAE and RD show some significant performance, especially in the localization task. Analogous to the ATLAS dataset, some feature-modeling methods (PaDiM and CFLOW-AD) fail. In contrast to the results on ATLAS, however, the image-reconstruction methods fail as well.
These observations are in line with our previous hypotheses: for image-reconstruction methods, localization performance strongly correlates with the contrast between lesions and healthy tissue, whereas feature-modeling methods are seemingly more capable of spotting abnormalities in morphology and texture, though struggle with small lesions. Among the self-supervised methods, PII and CutPaste achieve comparably high detection scores, but localize poorly, a common hint for confounding variables.
\begin{figure*}[!t]
\centerline{\includegraphics[width=\textwidth]{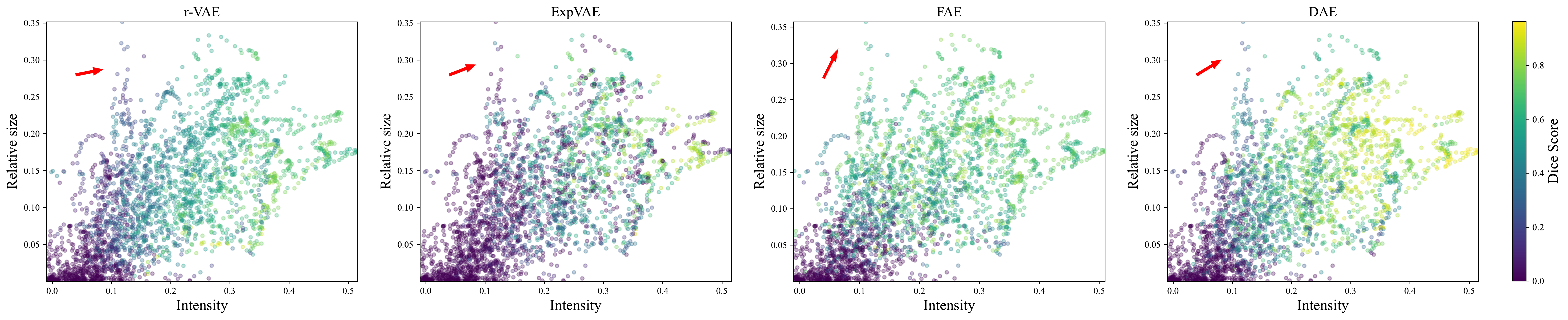}}
\caption{Pixel-wise Dice scores of r-VAE, ExpVAE, FAE, and DAE for anomalies of different size and intensity in BraTS-T2 images. The red arrows indicate the relative influence of each axis on the Dice score.}
\label{fig:scatter}
\end{figure*}
\smallskip\newline
\begin{table}[htb]
\centering
\caption{Image-level anomaly detection performance of the compared methods on the CheXpert dataset.}
\label{tab:cxr}
\resizebox{\linewidth}{!}{
\begin{tabular}{clcccccccc} 
\toprule
&& \multicolumn{2}{c}{\textbf{Pleural Effusion}} &  & \multicolumn{2}{c}{\textbf{Opacity}} &  & \multicolumn{2}{c}{\textbf{Enlarged Card.}}  \\ 
\cmidrule(r{4pt}){1-2}\cmidrule{3-4}\cmidrule{6-7}\cmidrule{9-10}
& Method & \api  & AUROC &  & \api  & AUROC &  & \api  & AUROC \\
\cmidrule(r{4pt}){1-2}\cmidrule{3-4}\cmidrule{6-7}\cmidrule{9-10}
\parbox[t]{2mm}{\multirow{4}{*}{\rotatebox[origin=c]{90}{\textbf{IR}}}}
& VAE & 0.55 & 0.57 &  & 0.51 & 0.54 &  & 0.56 & 0.56 \\
& r-VAE & 0.52 & 0.57 &  & 0.47 & 0.50 &  & 0.52 & 0.56 \\
& f-AnoGAN & 0.59 & 0.61 &  & 0.49 & 0.51 &  & 0.57 & 0.59 \\
& H-TAE-S &0.51& 0.51 &  & 0.50 & 0.50 &  & 0.53 & 0.51\\
\cmidrule(r{4pt}){1-2}\cmidrule{3-4}\cmidrule{6-7}\cmidrule{9-10}
\parbox[t]{2mm}{\multirow{4}{*}{\rotatebox[origin=c]{90}{\textbf{FM}}}}
& FAE & \textbf{0.73} & \textbf{0.77} &  & \textbf{0.57} & \textbf{0.61} &  & \textbf{0.63} & \textbf{0.66} \\
& PaDiM & 0.61 & 0.66 &  & 0.50 & 0.52 &  & 0.60 & 0.63 \\
& CFLOW-AD & 0.67 & 0.73 &  & 0.52 & 0.54 &  & 0.62 & \textbf{0.66} \\
& RD & 0.65 & 0.71 &  & 0.54 & 0.58 &  & 0.61 & 0.63 \\
\cmidrule(r{4pt}){1-2}\cmidrule{3-4}\cmidrule{6-7}\cmidrule{9-10}
\parbox[t]{2mm}{\multirow{2}{*}{\rotatebox[origin=c]{90}{\textbf{AB}}}}
& ExpVAE & 0.51 & 0.52 &  & 0.52 & 0.51 &  & 0.52 & 0.52 \\
& AMCons & 0.39 & 0.31 &  & 0.49 & 0.48 &  & 0.41 & 0.37 \\
\cmidrule(r{4pt}){1-2}\cmidrule{3-4}\cmidrule{6-7}\cmidrule{9-10}
\parbox[t]{2mm}{\multirow{3}{*}{\rotatebox[origin=c]{90}{\textbf{S-S}}}}
& PII & 0.64 & 0.65 &  & 0.53 & 0.54 &  & 0.56 & 0.56 \\
& DAE & 0.63 & 0.65 &  & 0.52 & 0.55 &  & 0.58 & 0.60 \\
& CutPaste & 0.58 & 0.59 &  & 0.53 & 0.54 &  & 0.57 & 0.58\\
\cmidrule(r{4pt}){1-2}\cmidrule{3-4}\cmidrule{6-7}\cmidrule{9-10}
& Random & 0.50 & 0.50 &  & 0.50 & 0.50 &  & 0.50 & 0.50 \\
\bottomrule
\end{tabular}}
\end{table}

\textbf{CheXpert.}
The ability of UPD methods to detect one of the three selected pathologies in CheXpert is generally low. The winners are - like in the other modalities - FAE and RD. Also, CFLOW-AD is among the better-performing methods in this evaluation. A clear finding here is that feature-modeling methods again outperform image-reconstruction ones in a low-contrast setting. Interestingly, pleural effusion gets consistently better detected by all models than opacity or enlarged cardiomediastinal contour.
%
%
\subsection{Summary of the Main Results}
Feature-modeling methods clearly dominate the leaderboards. For every but the image-level task on ATLAS, there is a feature-modeling method that surpasses the best method from all other groups. The clear outlier performances on brain-MRI, RF, and CXR come from FAE and RD, with FAE having a slight edge on the bigger lesions of BraTS, and RD being the best at segmenting the smaller lesions of ATLAS and DDR.

\subsection{Influence of Size and Intensity}\label{sec:influence_of_size_and_intensity}
In this section, we evaluate the sensitivity of representative models to differences in intensity and size of the anomalies on BraTS-T2.
We compute anomaly segmentations (binarized at the optimal operating point given by $\lceil$Dice$\rceil$ in Table \ref{tab:main}) from r-VAE, ExpVAE, FAE, and DAE for images with different lesion sizes (relative to the brain area) and intensities (compared to the average intensity in the slice).
The scatter-plots in Figure \ref{fig:scatter} show the pixel-wise Dice scores of each image and the relative influence of size and intensity on the performance of each model, indicated by red arrows.
Here, the individual influence of size and intensity of the lesions on their detection performance by different models can be observed.

As expected from the previous experiments, all methods struggle with small and non-hyperintense anomalies (bottom left corner).
Whereas r-VAE, ExpVAE, and DAE are strongly influenced by lesion intensity, the performance of FAE is less dependent on it, which is consistent with our findings and hypotheses from Section \ref{subsec:main_results}.
Interestingly, DAE performs exceptionally well for large and hyperintense anomalies, but the good performance is focused mostly on that region (with some outliers). This behavior mirrors the results of DAE from Section \ref{subsec:main_results} and highlights how a pre-text task with strong priors can lead to SOTA results, but at the cost of generalizability and versatility.
Moreover, while VAE is highly sensitive to the relative intensities of anomalies, it seems largely invariant to their size, confirming the findings from Section \ref{subsec:main_results}.
Finally, apart from showcasing the worst detection performance among the 4 models, ExpVAE achieves very low Dice scores even for some large, hyperintense anomalies, suggesting that the gradient w.r.t. the KL-divergence is not a reliable measure for detecting anomalies.

\subsection{Qualitative Results}\label{sec:qualitative_results}

\begin{figure*}[hbt]
\centerline{\includegraphics[width=19cm]{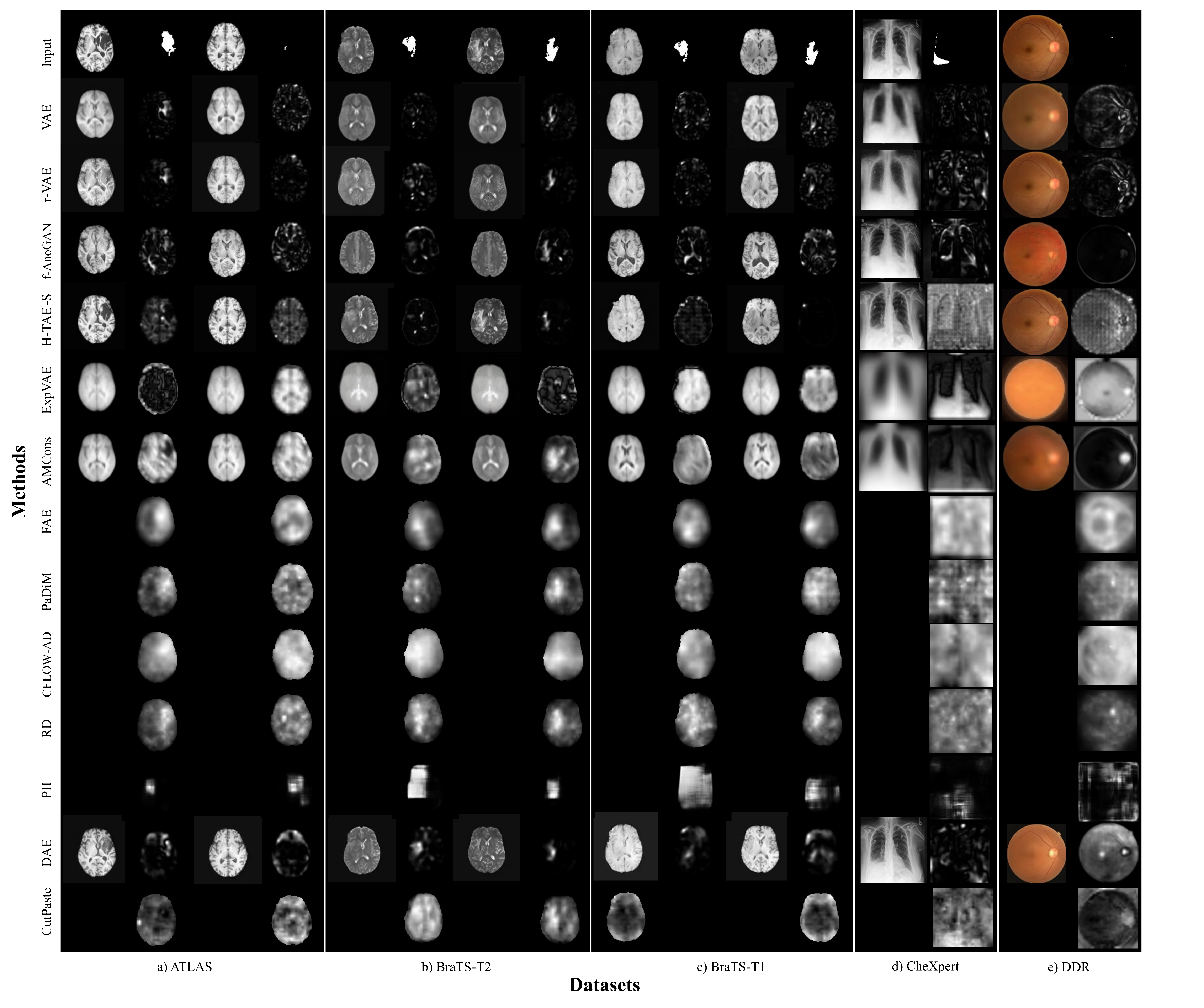}} 
\caption{Examples of model outputs (odd columns) and anomaly maps (even columns) of the reviewed methods for each modality. The first row includes the inputs, the corresponding lesion segmentations for BraTS and DDR, and one of the two Pleural Effusion ROI segmentations provided by the pixel level annotated subset of CheXpert \cite{chexplanation}.}
\label{fig2}
\end{figure*}

Uncurated, qualitative results of anomaly maps (and reconstructions where available) are shown in Fig. \ref{fig2}. It is clearly noticeable, that the anomaly maps of image-reconstruction models rarely look useful. This is in stark contrast to the feature-modeling methods that - although being coarse - often show higher anomaly values for anomalous regions. The maps of FAE and RD seem to be most localized, which is in accordance with their good overall performance. 
From the outputs of DAE, it can be seen that the model fails to faithfully remove anomalies. Considering also PII and the quantitative results of the two methods, we conclude that self-supervised learning with synthetic anomalies assumes strong priors and rarely generalizes well across different anomalies.
In the anomaly maps of AMCons, which are the activations from the first convolutional layer, we can observe amplification of high- and low-intensity regions from the input images. This poses difficulties if anomalies are not hyper-intense and explains the low performance of the method in all datasets but BraTS-T2.

The anomaly maps generated to localize the sample of pleural effusion on CheXpert show no useful information, which opens questions, since pleural effusion is generally a local pathology, and the quantitative performance - especially of FAE - is noticeable.
We can not rule out that the detection performance is considerable because of images with very different positions, anatomies, and even severe image corruptions in the part of the test set that is labeled with pleural effusion.  
Out-of-distribution samples are also considered anomalous by UAD methods, even when the anomalies of interest are only pathologies.
Meissen \textit{et al.} recently raised awareness to this problem in \cite{meissen_hyperkvasir}.
The DDR sample shows another surprising finding.
Here, the feature-modeling methods (especially RD) manage to localize the extremely small anomaly correctly.

\subsection{Self-Supervised Pre-training}
In this experiment, we investigate the effect of domain-specific self-supervised pre-training on the reviewed methods.
To do so, we pre-train the encoder parts of all reviewed methods using CCD \cite{ccd}, which combines the powerful contrastive loss of SimCLR \cite{simclr} with two self-supervised transformation prediction tasks to learn domain-specific representations in an unsupervised manner.
We hereby differentiate two distinct cases: For image-reconstruction, attention-based, and self-supervised anomaly detection methods, we investigate the effects of initialization with pre-trained over random features.\footnote{Note that H-TAE-S uses a Swin Transformer-type architecture and is thus not easily compatible with CCD.}.
For feature-modeling methods that leverage a frozen encoder (FAE, PaDiM, CFLOW-AD, and RD), we are interested to examine whether domain-specific representations created in a self-supervised manner are more effective for UPD performance than general purpose ones, created by pre-training on big natural image datasets. The effects of pre-training can be seen in Table \ref{tab:pretrain}.
\smallskip\newline

\renewcommand{\row}[5]{#1 & #4 && #3 && #2 && #5}
\begin{table*}
\centering
\caption{Detection and localization results of the image-reconstruction (\textbf{IR}), feature-modeling (\textbf{FM}), attention-based (\textbf{AB}), and self-supervised anomaly detection (\textbf{S-S}) methods on the brain-MRI and RF datasets after self-supervised pre-training.
The small numbers show the change compared to the results without self-supervised pre-training and are color-coded (green and red) if the change is statistically significant with $p < 0.05$ using a two-sided Welch's t-test over multiple random seeds.
}
\label{tab:pretrain}
\resizebox{\linewidth}{!}{
    \begin{tabular}{clccccccccccccccccccc}
        \toprule
        & \row{}{\multicolumn{4}{c}{\textbf{ATLAS}}}{\multicolumn{4}{c}{\textbf{BraTS-T1}}}{\multicolumn{4}{c}{\textbf{BraTS-T2}}}{ \multicolumn{4}{c}{\textbf{DDR}}}\\
        \cmidrule(r{4pt}){1-2}\cmidrule{3-6}\cmidrule{8-11}\cmidrule{13-16}\cmidrule{18-21}
        & & \multicolumn{2}{c}{image-level} & \multicolumn{2}{c}{pixel-level} && \multicolumn{2}{c}{image-level} & \multicolumn{2}{c}{pixel-level} && \multicolumn{2}{c}{image-level} & \multicolumn{2}{c}{pixel-level} && \multicolumn{2}{c}{image-level} & \multicolumn{2}{c}{pixel-level} \\[1mm]
        & Method & \api & AUROC & \app & \dice && \api & AUROC & \app & \dice && \api & AUROC & \app & \dice && \api & AUROC & \app & \dice \\
        \cmidrule(r{4pt}){1-2}\cmidrule{3-6}\cmidrule{8-11}\cmidrule{13-16}\cmidrule{18-21}
        \parbox[t]{2mm}{\multirow{5}{*}{\rotatebox[origin=c]{90}{\textbf{IR}}}}
        & \row{VAE}{0.57 \tu{} & 0.76 \tu{} & \textbf{0.11} \tu{} & 0.20 \tu{}}{0.64 \tu{} & 0.70 \tu{} & 0.13 \tu{} & 0.19 \tu{}}{0.68 \tu{} & 0.73 \tu{} & 0.27 \tn{.01} & 0.33 \tu{}}{0.48 \tu{} & 0.48 \tu{} & 0.02 \tu{} & 0.05 \tu{}} \\
        
        & \row{r-VAE}{\textbf{0.61} \tp{.01} & \textbf{0.79} \tp{.01} & 0.09 \tu{} & 0.17 \tu{}}{0.70 \tu{} & 0.76 \tu{} & 0.13 \tu{} & 0.19 \tu{}}{0.75 \tu{} & 0.77 \tu{} & 0.36 \tu{} & 0.40 \tu{}}{0.49 \tn{.03} & 0.48 \tn{.02} & 0.02 \tn{.01} & 0.09 \tu{}} \\
        
        & \row{f-AnoGAN}{0.28 \tp{.02} & 0.50 \tp{.04} & 0.02 \tu{} & 0.04 \tn{.02}}{0.48 \tu{} & 0.53 \tu{} & 0.06 \tu{} & 0.11 \tn{.01}}{0.63 \tp{.07} & 0.70  \tp{.09} & 0.15 \tu{} & 0.21 \tu{}}{0.41 \tn{.03} & 0.42 \tsign{.03} & 0.01 \tu{} & 0.01 \tu{}} \\
        
        & \row{H-TAE-S}{N/A & N/A & N/A & N/A}{N/A & N/A & N/A & N/A}{N/A & N/A & N/A & N/A}{N/A & N/A & N/A & N/A} \\
        \cmidrule(r{4pt}){1-2}\cmidrule{3-6}\cmidrule{8-11}\cmidrule{13-16}\cmidrule{18-21}
        
        \parbox[t]{2mm}{\multirow{4}{*}{\rotatebox[origin=c]{90}{\textbf{FM}}}}
        
        & \row{FAE}{0.53 \tsigp{.03} & 0.75 \tsigp{.02} & 0.10 \tsigp{.02} & \textbf{0.21} \tsigp{.03}}{\textbf{0.85} \tn{.01} & \textbf{0.85} \tu{} & \textbf{0.43} \tp{.01} & \textbf{0.46} \tp{.01}}{\textbf{0.89} \tsigp{.02} & \textbf{0.88} \tp{.01} &\textbf{0.55} \tsigp{.04} & \textbf{0.55} \tsigp{.03}}{0.55 \tsign{.09} & 0.57 \tsign{.06} & 0.03 \tsign{.04} & 0.08 \tsign{.07}} \\
        
        & \row{PaDiM}{0.41 \tp{.07} & 0.65 \tp{.09} & 0.07 \tp{.02} & 0.16 \tp{.03}}{0.74 \tsigp{.14} & 0.78 \tsigp{.13} & 0.26 \tp{.05} & 0.32 \tp{.04}}{0.81 \tsigp{.15} & 0.81 \tsigp{.13} & 0.47 \tsigp{.14} & 0.48 \tsigp{.10}}{0.54 \tn{.01} & 0.54 \tn{.01} & 0.02 \tu{} & 0.05 \tsign{.02}} \\
        
        & \row{CFLOW-AD}{0.45 \tsigp{.05} & 0.68 \tsigp{.06} & 0.06 \tsigp{.02} & 0.14 \tp{.04}}{0.76 \tp{.11} & 0.78 \tp{.09} & 0.24 \tsigp{.08} & 0.30 \tp{.06}}{0.79 \tsigp{.08} & 0.80 \tsigp{.08} & 0.31 \tu{} & 0.38 \tp{.03}}{0.53 \tp{.02} & 0.54 \tsigp{.03} & 0.02 \tn{.01} & 0.07 \tn{.01}} \\
        
        & \row{RD}{0.42 \tn{.13} & 0.64 \tn{.13} & 0.05 \tn{.06} & 0.09 \tn{.13}}{0.74 \tn{.07} & 0.78 \tsign{.05} & 0.28 \tsign{.08} & 0.35 \tsign{.07}}{0.84 \tn{.01} & 0.85 \tu{} & 0.39 \tn{.08} & 0.44 \tn{.06}}{0.53 \tsign{.13} & 0.54 \tsign{.10} & \textbf{0.04} \tsign{.06} & 0.09 \tsign{.10}} \\
    
        \cmidrule(r{4pt}){1-2}\cmidrule{3-6}\cmidrule{8-11}\cmidrule{13-16}\cmidrule{18-21}
        
        \parbox[t]{2mm}{\multirow{2}{*}{\rotatebox[origin=c]{90}{\textbf{AB}}}}
        & \row{ExpVAE} {0.40 \tp{.03} & 0.62 \tp{.05} & 0.01 \tu{} & 0.03 \tu{}}{0.58 \tp{.02} & 0.61 \tp{.05} & 0.05 \tsign{.02} & 0.12 \tn{.01}}{0.66 \tp{.03} & 0.68 \tp{.02} & 0.15 \tp{.03} & 0.22 \tp{.04}}{0.52 \tn{.01} & 0.53 \tn{.01} & 0.004 \tu{} & 0.01 \tu{}}  \\
        
        & \row{AMCons}{0.31 \tn{.01} & 0.52 \tn{.01} & 0.01 \tu{} & 0.03 \tu{}}{0.61 \tu{} & 0.64 \tu{} & 0.06 \tp{.01} & 0.12 \tu{}}{0.73 \tn{.05} & 0.74 \tn{.04} & 0.27 \tn{.08} & 0.34 \tn{.06}}{0.49 \tu{} & 0.50 \tp{.01} & 0.004 \tu{} & 0.01 \tu{}} \\
        
        \cmidrule(r{4pt}){1-2}\cmidrule{3-6}\cmidrule{8-11}\cmidrule{13-16}\cmidrule{18-21}
        \parbox[t]{2mm}{\multirow{3}{*}{\rotatebox[origin=c]{90}{\textbf{S-S}}}}
        
        & \row{PII}{0.32 \tn{.05} & 0.52 \tn{.08} & 0.02 \tn{.01} & 0.05 \tn{.02}}{0.46 \tn{.08} & 0.50 \tn{.14} & 0.07 \tn{.06} & 0.14 \tsign{.08}}{0.60 \tp{.03} & 0.66 \tp{.04} & 0.11 \tn{.02} & 0.19 \tn{.03}}{0.60 \tn{.02} & 0.60 \tn{.03} & 0.01 \tu{} & 0.04 \tp{.03}} \\
        
        & \row{DAE}{0.47 \tp{.03} & 0.68 \tp{.03} & 0.07 \tp{.02} & 0.15 \tp{.02}}{0.71 \tp{.01} & 0.75 \tp{.01} & 0.14 \tp{.01} & 0.20 \tu{}}{0.81 \tu{} & 0.81 \tp{.01} & 0.48 \tp{.01} & 0.50 \tp{.01}}{0.59 \tp{.05} & 0.60 \tp{.05} & 0.01 \tu{} & 0.03 \tu{}} \\
        
        & \row{CutPaste} {0.49 \tp{.11} & 0.71 \tsigp{.13} & 0.02 \tn{.01} & 0.04 \tn{.02}}{0.68 \tp{.07} & 0.73 \tp{.08} & 0.07 \tu{} & 0.13 \tu{}}{0.66 \tsigp{.07} & 0.70 \tp{.07} & 0.31 \tp{.09} & 0.35 \tp{.09}}{\textbf{0.70} \tp{.06} & \textbf{0.65} \tp{.05} & \textbf{0.04} \tsigp{.02} & \textbf{0.11} \tsigp{.05}} \\
        
        \cmidrule(r{4pt}){1-2}\cmidrule{3-6}\cmidrule{8-11}\cmidrule{13-16}\cmidrule{18-21}
        & \row{Random}{0.30  & 0.50  & 0.02  & 0.03}{0.48  & 0.50  & 0.06  & 0.11}{0.48  & 0.50  & 0.06  & 0.11}{0.50  & 0.50  & 0.004  & 0.01 } \\
        \bottomrule
    \end{tabular}}
\end{table*}

\textbf{Weight Initialization.}
We notice no significant benefits to performance for any investigated method except CutPaste.
These results are consistent with Raghu \textit{et al.}\cite{transfusion}, in that weight transfer offers marginal to no benefits over random initialization on medical datasets, and we extend their results to the unsupervised setting using domain-specific weights created with self-supervision.
Interestingly thought, CutPaste benefits from pre-training and even becomes the best-performing model in image-level detection on DDR. We hypothesize that this exception may be rooted in the similarity between CutPaste and CCD in their learning approach.
\smallskip\newline
\textbf{Backbone pre-training.}
For all brain-MRI datasets, we notice increased performance on every feature-modeling method but RD, with PaDiM benefiting the most as they approach segmentation performance close to the best-performing models without pre-training on BraTS-T2.
For CheXpert, performance is mostly worse with some exceptions where we notice similar or marginally better performance, while for DDR, the effect of pre-training is consistently negative, with a single exception for CFLOW-AD.
An interesting outlier in this class of methods is RD which declines strongly in performance after pre-training.
We assume that because anomaly detection with distillation models relies on differences between the teacher and the student, the representations of a teacher that was trained on the domain of interest might be easier to match by the student, and, thus, important differences might be missed.
We hypothesize that while domain-specific, self-supervised pre-training can create useful, application-specific features that frozen backbone methods can benefit from, the usefulness of these representations for the downstream tasks is highly dependent on the pretext task and might vary strongly for different modalities, as well as anomalies. Still, the results on BraTS showcase how the inherent ability of feature-modeling methods to benefit from better representations can significantly alter evaluation results, emphasizing the importance of studying self-supervised pre-training algorithms in conjunction with UAD methods in feature works.
\begin{table}[t]
\centering
\caption{Image-level anomaly detection performance of the compared methods on CheXpert after self-supervised pre-training.
The small numbers show the change compared to the results without self-supervised pre-training and are color-coded (green and red) if the change is statistically significant with $p < 0.05$ using a two-sided Welch's t-test over multiple random seeds.
}
\label{tab:cxr_ccd}
\resizebox{\linewidth}{!}{
    \begin{tabular}{clcccccccc} 
    \toprule
    \multicolumn{2}{c}{} & \multicolumn{2}{c}{\textbf{Pleural Effusion}} &  & \multicolumn{2}{c}{\textbf{\textbf{Opacity}}} &  & \multicolumn{2}{c}{\textbf{Enlarged Card.}}  \\ 
    \cmidrule(r{4pt}){1-2}\cmidrule{3-4}\cmidrule{6-7}\cmidrule{9-10}
    & Method & \api  & AUROC &  & \api  & AUROC &  & \api  & AUROC \\ 
    \cmidrule(r{4pt}){1-2}\cmidrule{3-4}\cmidrule{6-7}\cmidrule{9-10}
    \parbox[t]{2mm}{\multirow{4}{*}{\rotatebox[origin=c]{90}{\textbf{IS}}}}
    & VAE & 0.56 \tp{.01} & 0.58 \tp{.01} &  & 0.52 \tp{.01} & 0.54 \tu{} &  & 0.56 \tu{} & 0.57 \tp{.01} \\
    & r-VAE & 0.52 \tu{} & 0.57 \tu{} &  & 0.47 \tu{} & 0.49 \tn{.01} &  & 0.52 \tu{} & 0.56 \tu{} \\
    & f-AnoGAN & 0.60 \tp{.01} & 0.62 \tp{.01} &  & 0.50 \tp{.01} & 0.52 \tp{.01} &  & 0.58 \tp{.01} & 0.60 \tp{.01} \\
    & H-TAE-S & N/A & N/A &  & N/A & N/A &  & N/A & N/A \\
    \cmidrule(r{4pt}){1-2}\cmidrule{3-4}\cmidrule{6-7}\cmidrule{9-10}
    \parbox[t]{2mm}{\multirow{4}{*}{\rotatebox[origin=c]{90}{\textbf{FS}}}}
    & FAE & \textbf{0.74} \tp{.01} & \textbf{0.78} \tp{.01} &  & \textbf{0.55} \tsign{.02} & \textbf{0.58} \tsign{.03} &  & \textbf{0.65} \tsigp{.02} & \textbf{0.68} \tsigp{.02} \\
    & PaDiM & 0.59 \tsign{.02} & 0.63 \tsign{.03} &  & 0.48 \tsign{.02} & 0.49 \tsign{.03} &  & 0.58 \tsign{.02} & 0.60 \tsign{.03} \\
    & CFLOW-AD & 0.58 \tsign{.09} & 0.62 \tsign{.11} &  & 0.47 \tsign{.05} & 0.48 \tsign{.06} &  & 0.58 \tsign{.04} & 0.60 \tsign{.06} \\
    & RD & 0.66 \tp{.01} & 0.69 \tn{.02} &  & \textbf{0.55} \tp{.01} & \textbf{0.58} \tu{} &  & 0.61 \tu{} & 0.62 \tn{.01} \\
    \cmidrule(r{4pt}){1-2}\cmidrule{3-4}\cmidrule{6-7}\cmidrule{9-10}
    \parbox[t]{2mm}{\multirow{2}{*}{\rotatebox[origin=c]{90}{\textbf{AB}}}}
    & ExpVAE & 0.48 \tn{.03} & 0.47 \tn{.05} &  & 0.51 \tn{.01} & 0.49 \tn{.02} &  & 0.49 \tn{.03} & 0.48 \tn{.04} \\
    & AMCons & 0.43 \tp{.04} & 0.40 \tp{.09} &  & 0.50 \tp{.01} & 0.52 \tp{.04} &  & 0.45 \tp{.04} & 0.43 \tp{.06} \\
    \cmidrule(r{4pt}){1-2}\cmidrule{3-4}\cmidrule{6-7}\cmidrule{9-10}
    \parbox[t]{2mm}{\multirow{3}{*}{\rotatebox[origin=c]{90}{\textbf{S-S}}}}
    & PII & 0.65 \tp{.01} & 0.66 \tp{.01} &  & 0.52 \tn{.01} & 0.53 \tn{.01} &  & 0.58 \tp{.02} & 0.59 \tp{.03} \\
    & DAE & 0.62 \tn{.01} & 0.64 \tn{.01} &  & 0.52 \tu{} & 0.54 \tn{.01} &  & 0.58 \tu{} & 0.60 \tu{} \\
    & CutPaste & 0.55 \tsign{.03} & 0.58 \tn{.01} &  & 0.52 \tn{.01} & 0.54 \tu{} &  & 0.55 \tn{.02} & 0.56 \tsign{.02} \\
    \cmidrule(r{4pt}){1-2}\cmidrule{3-4}\cmidrule{6-7}\cmidrule{9-10}
    & Random & 0.50 & 0.50 &  & 0.50 & 0.50 &  & 0.50 & 0.50 \\
    \bottomrule
    \end{tabular}}
\end{table}

\subsection{Complexity Analysis}
Fast diagnosis is critical for several clinical applications, e.g. in the emergency department.
Moreover, real-time diagnostic procedures like Colonoscopy require algorithms capable of online evaluation.
Therefore, a focus has been placed on the development of methods with fast inference \cite{fanogan}, \cite{fastdiffusion}.
In Table \ref{tab:time}, we report inference times for all reviewed methods.
Results are created using a single Nvidia Quadro RTX 8000 GPU and an AMD Ryzen Threadripper 3960X CPU. As timing measurements are very machine/setup dependant, results can vary and are presented here only for qualitative comparisons. We notice that even though r-VAE and PaDiM can show strong performance on the MR datasets, they are significantly slower than the other methods.

\begin{table}[t]
\centering
\caption{Inference speed in frames per second and size complexity in million parameters for all selected methods.
\label{tab:time}}
\resizebox{\linewidth}{!}{
\begin{tabular}{lcclcc}
\cmidrule(r{4pt}){1-3}\cmidrule(r{4pt}){4-6}
Method   & \begin{scriptsize}Speed [fps]\end{scriptsize}  & \begin{scriptsize}Params (M)\end{scriptsize} &Method   & \begin{scriptsize}Speed [fps]\end{scriptsize}  & \begin{scriptsize}Params (M)\end{scriptsize} \\ 
\cmidrule(r{4pt}){1-3}\cmidrule(r{4pt}){4-6}
VAE  & 443.8  & 3&PaDiM  & 0.5 & 69 \\
r-VAE  & 0.4 & 3 & CFLOW-AD  & 72.9 & 237 \\
f-AnoGAN  & 47.0 & 28 & RD  & 109.7 & 161 \\
H-TAE-S & 131.3 & 47 & PII & 568.8 & 3 \\
AMCons & 265.9 & 48& CutPaste & 398.9 & 27 \\
ExpVAE & 194.0 & 269 & DAE & 241.9 & 276 \\
FAE & 191.5 & 21 &   &   &  \\
\cmidrule(r{4pt}){1-3}\cmidrule(r{4pt}){4-6}
\end{tabular}}
\end{table}

\begin{figure*}[!t]
\centerline{\includegraphics[width=\textwidth]{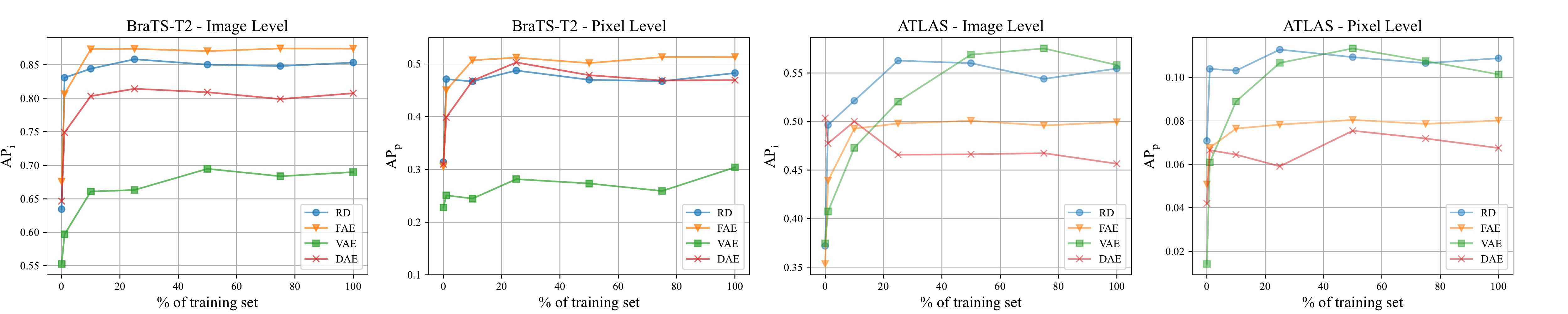}}
\end{figure*}
\begin{figure*}[!t]
\centerline{\includegraphics[width=0.78\textwidth]{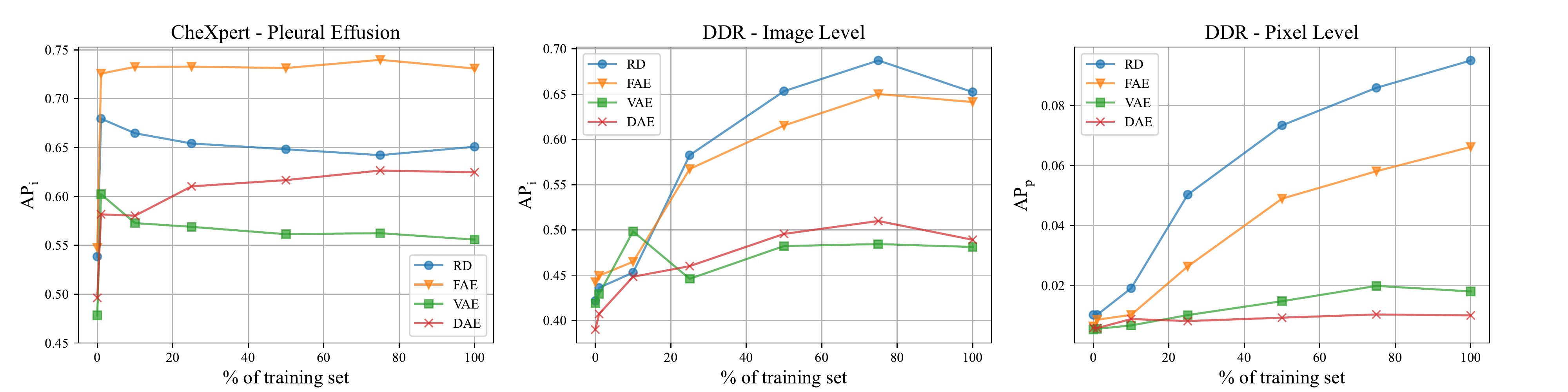}}
\caption{Detection and localization results of RD, FAE, VAE, and DAE when trained on different fractions of the training set.}
\label{fig:percentage}
\end{figure*}
We also report the number of parameters as a measure of space complexity.
As expected, RD, CFLOW-AD, and to a lower extent PaDiM, are notably more resource-intensive, since they are implemented with a Wide ResNet-50-2 backbone \cite{wideresnet}.  
While this can limit their usability with mobile or older systems, feature-modeling methods enable a trade-off between complexity and UPD performance, as their backbones can be easily exchanged.
Also, pruning of the backbone can be considered here to considerably save computation power.


\subsection{The Effects of Limited Training Data}\label{sec:the_effects_of_limited_training_data}
We further study the model performance when utilizing subsets of each training dataset. In Fig. \ref{fig:percentage}, we plot image- and pixel-level AP of FAE, RD, DAE and VAE when trained with $x$\% of each modality's training set $\forall x\in\{0,1,5,10,25,50,75,100\}$, with $x=0$ corresponding to just a single image or MRI volume. 

On all datasets but DDR, we observe a -- usually steep -- increase in performance from single image/volume to 1\% and at around 10 to 25\%, performance generally plateaus, except for VAE on ATLAS, where performance increases until 50 to 75\%.
Interestingly, all three other methods are on par or outperform VAE on the segmentation task of BraTS-T2 even when trained on a single volume. 
This is an interesting behavior that brings into question the ability of anomaly detection models to effectively utilize large amounts of data, one of the main advantages of UPD as described in Section \ref{sec:introduction}. On the other hand, it might also point towards an increased data-efficiency for these models. In any case, it requires further investigation in the future.
Finally, models provided with an increasing number of DDR train samples demonstrate a behavior closer to the theoretically expected one. For Image-level detection, performance slowly increases as we allow more train samples and saturates closer to $x=100\%$.  For segmentation, performance seemingly still tends to increase at $x=100\%$, suggesting a bottleneck in the number of available train samples.


\section{Discussion and Limitations}
Our experiments have shown that feature-modeling methods significantly outperform methods from the other categories evaluated in this work.
They have proven to be more generalizable to different types of anomalies that appear in medical images and are less reliant on intensity differences between normal and abnormal tissue.
However, we have also identified a weakness of feature-modeling methods with regard to small anomalies.
Our experiments have further revealed that the strong priors of methods that use artificial anomalies and pretext tasks hinder their generalization to anomalies with even slightly different characteristics than those they have been designed for.

Another interesting finding from our experiments is that good performance on the MVTec-AD dataset \cite{MVTecDataset} does not necessarily translate to good performance on medical images.
While RD is clearly among the best-performing models in our study, PaDiM, CFLOW-AD, and CutPaste perform rather poorly.
Given the diversity in the approaches, a common design choice for this performance drop is difficult to pin down.
The classes in MVTec-AD, however, have simpler structures than the medical images considered in our study and exhibit lower inter-sample variance, making the normative distribution easier to learn, such that good anomaly detection performance is easier to achieve.

Although the results of some methods are impressive -- especially on brain MRI -- the overall performance is far from being clinically useful yet.
Moreover, while great care has been taken in both the selection of datasets, and optimization of methods, we can neither guarantee that the datasets are representative of the domain of medical images nor that all methods have reached their optimal performance.
Especially some of the well-performing models on industrial defect detection mentioned above underperform in our study.
While our conclusion for this finding lies in the differences between these datasets, the risk that certain models have not reached their optimal hyperparameter configuration cannot be excluded.
On the other end, FAE was developed by multiple authors of this manuscript and is thus likely to have reached its optimal performance.

\section{Conclusion}
This work has thoroughly evaluated the current state of the art in unsupervised pathology detection in medical images.
It has not only answered questions and put various approaches into perspective, but it has also opened new questions to be researched in works to come: First of all, feature-modeling methods are currently under-explored in the medical UAD domain yet have shown great potential in our evaluations. In this context, domain-specific pre-training is also under-explored. While we have seen increased performance in some datasets through this technique, it was harmful to others. Further research must therefore be conducted to investigate the effectiveness of different pretext tasks and pre-training techniques for UPD.
So far, none of the methods considered in this study considered the context-dependent notion of normality. What might be regarded as anomalous for a young adult might still be in the realm of normality for an elderly person. We consider this a valuable future venue with great potential for increased performance.
Lastly, our work has touched upon a topic with great ethical implications: Currently, the saturation of performance after only a few samples (Section \ref{sec:the_effects_of_limited_training_data}) is preventing the effective use of large datasets. 
The same dilemma will cause models to skip underrepresented variations if the bottleneck is too small. Since the distribution of minorities in a society is inherently underrepresented, UPD is expected to work worse for minorities. In the context of this study, we want to raise awareness of this problem and urge researchers to develop solutions to mitigate it.

\bibliographystyle{IEEEtran}
\bibliography{bibliography}
\end{document}